\documentclass[11pt]{article}
\usepackage{lmodern}
\PassOptionsToPackage{numbers, sort&compress}{natbib}
\usepackage[margin=0.8in]{geometry}
\usepackage[utf8]{inputenc} % allow utf-8 input
\usepackage[T1]{fontenc}    % use 8-bit T1 fonts

\RequirePackage[colorlinks,citecolor=blue,urlcolor=blue]{hyperref}

\usepackage{url}            % simple URL typesetting
\usepackage{booktabs}       % professional-quality tables
\usepackage{amsfonts}       % blackboard math symbols
\usepackage{nicefrac}
\usepackage{wrapfig}% compact symbols for 1/2, etc.
\usepackage{microtype}      % microtypography
\usepackage{amsmath, amsthm} % tr`es bon mode math?matique
\usepackage{amsfonts,amssymb}
\usepackage{nicefrac}       % compact symbols for 1/2, etc.
\usepackage{microtype}      % microtypography

\usepackage{graphicx}
\usepackage{wrapfig,lipsum}
\usepackage[font=small,labelfont=bf]{caption}
\usepackage{subcaption}
\usepackage{thmtools, thm-restate}
\usepackage{mathrsfs} %pour mathscr

\usepackage[shortlabels]{enumitem}

\usepackage{cite}

\usepackage[capitalize]{cleveref}

\usepackage[utf8]{inputenc} % allow utf-8 input
\usepackage[T1]{fontenc}    % use 8-bit T1 fonts
\usepackage{hyperref}       %

\usepackage{url}            % simple URL typesetting
\usepackage{booktabs}       % professional-quality tables
\usepackage{amsfonts}       % blackboard math symbols
\usepackage{nicefrac}       % compact symbols for 1/2, etc.
\usepackage{microtype}      % microtypography
\usepackage{textcomp}
\usepackage{multirow}
\usepackage[table,xcdraw]{xcolor}

\usepackage{amsmath,amssymb}
\usepackage{amscd}

\usepackage{thmtools, thm-restate}
\usepackage{mathrsfs} %pour mathscr
\usepackage[usenames,dvipsnames]{pstricks}

\usepackage{graphicx}
\usepackage[capitalize]{cleveref}

\usepackage{algorithmic}
\usepackage[ruled,vlined]{algorithm2e}

\usepackage{color}
\definecolor{darkgreen}{rgb}{0,0.5,0}
\definecolor{purple}{rgb}{1,0,1}
\newcommand{\kibitz}[2]{\ifnum\Comments=1\textcolor{#1}{#2}\fi}
\definecolor{dgreen}{rgb}{0.00,0.49,0.00}
\definecolor{dblue}{rgb}{0,0.08,0.75}
\hypersetup{
    colorlinks = true,
    citecolor=dgreen,
    urlcolor=dblue,
    linkcolor=dblue
}

\renewcommand{\paragraph}[1]{\vspace{1em}\noindent{\bfseries #1}.}

\definecolor{dgreen}{rgb}{0.00,0.49,0.00}
\definecolor{dblue}{rgb}{0,0.08,0.75}

\crefname{assumption}{Assumption}{Assumptions}
\crefname{equation}{}{}
\crefname{figure}{Fig.}{Figs.}
\crefname{table}{Tab.}{Tabs.}
\crefname{section}{Sec.}{Sec.}
\crefname{theorem}{Thm.}{Thm.}
\crefname{lemma}{Lemma}{Lemmas}
\crefname{corollary}{Cor.}{Cor.}
\crefname{example}{Example}{Examples}
\crefname{remark}{Remark}{Remarks}
\crefname{algorithm}{Alg.}{Algorightms}
\crefname{algocf}{Alg.}{Algorightms}

\crefname{appendix}{Appendix}{Appendices}
\crefname{subappendix}{Appendix}{Appendices}
\crefname{subsubappendix}{Appendix}{Appendices}

\newcommand{\loss}{{\bigtriangleup}}
\newcommand{\R}{{\mathbb{R}}}

\newcommand{\argmin}{\operatornamewithlimits{argmin}}
\newcommand{\argmax}{\operatornamewithlimits{argmax}}

\newcommand{\X}{{\mathcal{X}}}
\newcommand{\Y}{{\mathcal{Y}}}

\newcommand{\D}{{\mathcal{D}}}

\newcommand{\fhat}{\widehat{f}}

\renewcommand{\paragraph}[1]{~\newline\noindent{\bfseries #1.}}

\newcommand{\eqal}[1]{\begin{align}#1\end{align}}

\newcommand{\crisp}{CRiSP}
\newcommand{\gt}{\Tilde{g}}

\newcounter{theoremdecente}[section]
\newenvironment{theoremdecente}[1][]
{\refstepcounter{theoremdecente}\par\medskip\noindent{\bfseries Theorem~\thetheoremdecente~ (#1).}\itshape}
{}

\crefname{theoremdecente}{Thm.}{Thm.}

\newenvironment{proofdecente}{\par\noindent{\itshape Proof.}}{\hfill$\square$\medskip}

\newcommand{\customtitle}{Structured Prediction for \crisp{} Inverse Kinematics \\Learning with Misspecified Robot Models}

\date{}

\author{Gian Maria Marconi$^{*,1,2}$ \quad  Raffaello Camoriano$^{*,2}$ \quad Lorenzo Rosasco$^{2,3,4}$ \quad Carlo Ciliberto$^{5}$ \\ {\footnotesize\em gianmaria.marconi@riken.jp}~~\qquad{\footnotesize\em raffaello.camoriano@iit.it}\qquad~~~~{\footnotesize\em lrosasco@mit.edu}\quad\qquad{\footnotesize\em c.ciliberto@ucl.ac.uk}~~}

\title{\LARGE\bf\customtitle{}}

\begin{document}

\maketitle

\begin{abstract}
With the recent advances in machine learning, problems that traditionally would require accurate modeling to be solved analytically can now be successfully approached with data-driven strategies. Among these, computing the inverse kinematics of a redundant robot arm poses a significant challenge due to the non-linear structure of the robot, the hard joint constraints and the non-invertible kinematics map. Moreover, most learning algorithms consider a completely data-driven approach, while often useful information on the structure of the robot is available and should be positively exploited.
In this work, we present a simple, yet effective, approach for learning the inverse kinematics. We introduce a structured prediction algorithm that combines a data-driven strategy with the model provided by a forward kinematics function -- even when this function is misspecified -- to accurately solve the problem. The proposed approach ensures that predicted joint configurations are well within the robot's constraints. We also provide statistical guarantees on the generalization properties of our estimator as well as an empirical evaluation of its performance on trajectory reconstruction tasks.
\end{abstract}
{\let\thefootnote\relax\footnote{{$^{*}$Equal Contribution. $^{1}$RIKEN Center for AI Project, Tokyo, Japan. $^{2}$IIT@MIT - Laboratory for Computational and Statistical Learning (LCSL), Istituto Italiano di Tecnologia, Genoa, Italy, and
Massachusetts Institute of Technology, Cambridge, MA, USA. $^{3}$MaLGa \& DIBRIS, Università degli Studi di Genova, Genova, Italy. $^{4}$Center for Brains, Minds and Machines, MIT, Cambridge, MA, USA. $^{5}$Computer Science, University College London, London, United Kingdom}}}
% \footnote{* Equal Contribution. $^{1}$RIKEN Center for AI Project, Tokyo, Japan. $^{2}$Laboratory for Computational and Statistical Learning, Istituto Italiano di Tecnologia, Italy, and
% Massachusetts Institute of Technology, Cambridge, MA, USA. $^{3}$MaLGa \& DIBRIS, Università degli Studi di Genova, Genova, Italy. $^{4}$Center for Brains, Minds and Machines, MIT, Cambridge, MA, USA. $^{5}$Computer Science, University College London, London, United Kingdom}
% \keywords{Structured prediction, inverse kinematics, machine learning} 
%===============================================================================

\section{Introduction}

Computing the inverse kinematics of a robot is a well-known key problem in several applications requiring robot control \cite{spong2006robot}. This task consists in finding a set of joint configurations that would result in a given pose of the end effector, and is traditionally solved by assuming access to an accurate model of the robot and employing geometric or numerical optimization techniques. However, a major drawback of these strategies is that they are extremely sensitive to inaccuracies in the model. 
This can be a significant limitation in settings where the kinematic parameters of the robot are only available up to a given precision.
A recently proposed alternative to model-based approaches is to learn the inverse kinematics function from examples of joint configurations and workspace pairs~\cite{d2001learning, de2008learning, oyama2001inverse, oyama2005inverse}. However, traditional regression techniques are not suited for this task since computing the inverse kinematics of robots with redundant joints is an ill-posed problem and there are multiple joint configurations that correspond to the same workspace pose~\cite{siciliano1990kinematic}. To address this issue, previous works have recast the problem in the velocity domain. The goal becomes that of learning a map from the velocity of the end effector to the velocity of joints. Alternatively, because this problem is locally linear~\cite{tevatia2000inverse}, regression techniques can be employed to compute piecewise functions. This idea has been explored both with linear estimators and neural networks (NN)~\cite{tevatia2000inverse,oyama2001inverse}. 
In contrast to learning in the velocity domain, works in \cite{jordan1992forward,oyama2005inverse,bocsi2011learning} proposed to solve the learning problem in the position domain. Albeit more challenging, this strategy has the advantage that it does not limit the algorithm to compute local inverse kinematics estimators but allows for more global solutions. 

\begin{figure}[t]
\centering
\includegraphics[width=0.4\textwidth]{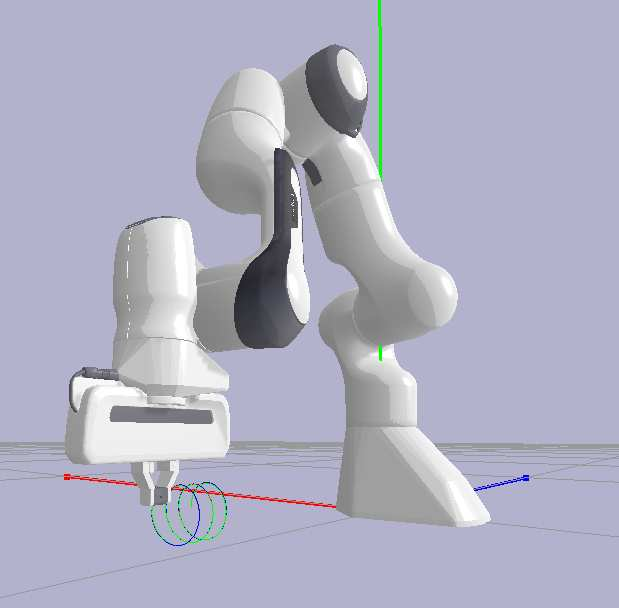}
\caption{Franka Emika Panda arm tracking a spiral trajectory via the \crisp{}-FK inverse kinematics structured predictor.}
\label{fig:panda-frontpage}
\end{figure}

In this work, we propose a novel structured prediction strategy to learn the inverse kinematics of a redundant robot. Our approach combines a data-driven strategy with the (possibly inaccurate or biased) forward kinematics function of the robot, potentially obtaining the best of both worlds.
We empirically show that our approach can compensate for biases in the forward kinematics and still learn an accurate inverse kinematics. This scenario is common when it is possible to gather data from a real robot, but the available kinematic model is imprecise. 
Our approach aims to estimate the joint configuration required to achieve a target pose, in contrast with most previous data-driven methods, which only consider the position of the end effector. 
We test our approach on trajectory reconstruction applications. 
As a byproduct of our work, we also provide a new dataset for inverse kinematics learning on a 5-DoF planar manipulator and on the 7-DoF Panda robot (see \cref{fig:panda-frontpage}). 
%
%
% % Plot panda joints circ

The remainder of this paper is organized as follows: in~\Cref{sec:ik_problem} we introduce the problem of inverse kinematics and review related work on the topic. In~\Cref{sec:struct_proposed} we introduce our method for inverse kinematics and characterize its theoretical properties. In~\Cref{sec:experiments} we empirically evaluate the proposed approach on trajectory reconstruction applications. \Cref{sec:conclusion} concludes this work and discusses potential directions for future research.

%===============================================================================

\section{Background and Problem Formulation}
\label{sec:ik_problem}
We introduce here the problem of learning the inverse kinematics of a robotic manipulator and discuss previous work on the topic. Let $SO(d)$ be the  special orthogonal group of dimension $d \in \mathbb{N}^+$. 
We denote by $\X = \R^d \times SO(d)$ 
the space of possible {\itshape end-effector poses}, comprising the Cartesian position and orientation of the robot's end effector in a $d$-dimensional Euclidean space. Assuming a robot with $J\in~\mathbb{N}^+$ joints, we denote by $\Y = [a_{1}, b_{1}] \times \ldots \times [a_{J}, b_{J}]$ the space of all {\itshape admissible joint configurations} such that for any $j~\in~\{1,\ldots,J\}$ the set  $[a_j,b_j]\subset [0,2\pi)$ identifies the physical limits of the $j$-th joint.

Assuming the robot to have forward kinematics $g:\Y\to\X$, our goal is to learn an inverse kinematics map, namely a function $f \colon \X \to \Y$ such that 
\begin{align}\label{eq:ik}
    g \circ f (x) = x.
\end{align}
However, finding such function is not straightforward. 
Since the forward kinematics $g$ of redundant manipulators is not injective, there are multiple joint configurations that result in the same end-effector pose. 
A common approach to this problem consists in defining it in the velocity domain and enforcing the uniqueness of the solution with further constraints. 
The resulting problem can be solved numerically.
However, the solution can be highly sensitive to 
%possible 
model inaccuracies (i.e., it needs very good knowledge of $g$)~\cite{george2018control}.

Data-driven approaches can overcome model inaccuracies by learning the inverse kinematics function from input-output pairs. 
Assuming $g$ to be {\itshape unknown}, these methods aim to learn $f$ by relying on a finite number $n$ of examples $(x_i,y_i)_{i=1}^n$ such that $y_i = g(x_i)$. To allow for a statistical analysis of our proposed method, in the rest of this work we will assume the pairs $(x_i,y_i)$ to be sampled i.i.d. according to a distribution $\rho$ on $\X\times\Y$. This can be done even when $g$ is not available by first randomly sampling a joint configuration $y_i$ and then measuring the robot pose $x_i = g(y_i)$ (see \cref{sec:experiments} for more details on this process in practice).

The work in~\cite{jordan1992forward} is an example of this. A NN learns the forward kinematics of the robot, and it is then used to train a second NN such that the composition of the two is the identity function. However, given the non-convex nature of optimization problems associated to NN models, this method is significantly unstable during training.
In~\cite{oyama2005inverse}, a method to learn a locally linear function in the velocity domain is proposed. 
This results in a good local approximation which, however, lacks global smoothness. 
The work in~\cite{bocsi2011learning} is related to our proposed approach in that it tackles the inverse kinematics problem directly in the Cartesian domain by using structured prediction techniques \cite{bakir2007predicting}.
By training a one-class structured SVM, their algorithm learns the inverse kinematics for some trajectories. However, this approach has a high sample complexity and requires to sample the training set only in a neighbourhood of the goal trajectory. 
This poses a challenge in most practical applications since it requires to retrain the model for each new trajectory or limit the usage of a model to trajectories close to the one used for training.

Similarly to \cite{bocsi2011learning}, in this work we rely on ideas from structured prediction to learn the inverse kinematics of a robot. However, we significantly improve over previous work by providing the following contributions: 
1) we introduce a novel approach that takes full advantage of useful side information, such as partial or inaccurate knowledge of the forward kinematics; 
2) we extend the learning problem to both position and orientation; 
3) we characterize the theoretical properties of the proposed estimator, proving universal consistency and excess risk bounds; 
4) we empirically demonstrate the effectiveness of our approach on a number of challenging and realistic scenarios in simulation. 

\begin{figure*}[htp]
  \centering
  \begin{subfigure}[t]{.45\linewidth}
    \centering
    \includegraphics[width=\textwidth]{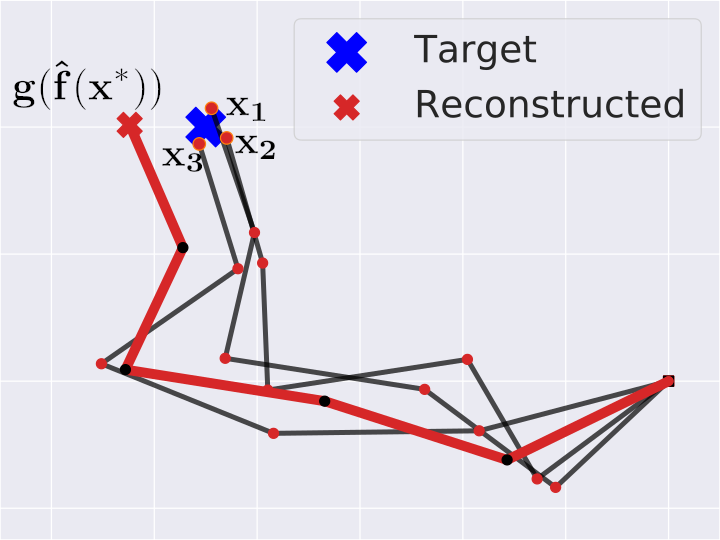}
    \caption{Configuration space loss}
    \label{fig:loss_joints}
  \end{subfigure}
 \qquad
  \centering
  \begin{subfigure}[t]{.45\linewidth}
    \centering
    \includegraphics[width=\textwidth]{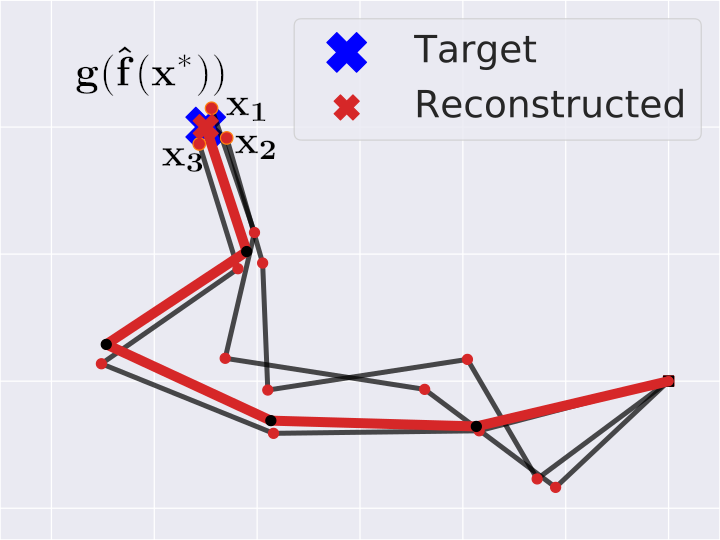}
    \caption{Forward kinematics loss}
    \label{fig:loss_cartesian}
  \end{subfigure}

  \caption{{\bfseries Dependency of learning inverse kinematics with respect to the structured prediction loss.} Pose reconstruction for $x^* = g(y^*)$ for the 5-DoF planar manipulator described in \cref{sec:robot-setup} given three training examples with pose $x_1, x_2, x_3$ close to the target pose $x^*$, but significantly different joint configurations $y_1,y_2,y_3$. Applying \crisp{} with the configuration space loss, produces configurations that are far from the target in Cartesian space (\cref{fig:loss_joints}). In contrast, the forward kinematic loss on \cref{eq:fk_loss} (even if possibly inaccurate) leads to a significantly better pose estimation (\cref{fig:loss_cartesian}).}
  \label{joint-vs-Cartesian}
\end{figure*}

\section{Structured Prediction for Inverse Kinematics}\label{sec:struct_proposed}

Structured prediction methods are used to learn input-output relations when the output space is not a linear space, but still presents some relevant structure that can be leveraged to compare and identify the best prediction for a given input. 
Notable examples are quantile estimation~\cite{takeuchi2006nonparametric}, image segmentation~\cite{nowozin2011structured}, manifold-valued prediction~\cite{rudi2018manifold}, and protein folding~\cite{joachims2009predicting}. 
Given an input $x \in \X$,  and a structured output $y \in \Y$,  many of the methods used in structured prediction can be abstracted as learning a function $F_{\alpha} \colon \X \times \Y \to \R$ that measures the quality of a candidate $y$ as a prediction for a specific input $x$ and depends on some learnable parameters $\alpha \in \Theta$. 
The structured estimator $\fhat \colon \X \to \Y$ is then the function identifying the best output that maximizes $F$

\begin{equation}\label{eq:struct_generic_estimator}
    \fhat(x) = \argmax_{y \in \Y}~ F_{\alpha}(y, x)
\end{equation}

While structured methods are often more computationally expensive than traditional supervised learning approaches, they can be applied to much more complex spaces and encode all the extra information that such spaces entail. 
In the case of inverse kinematics learning, the output space consisting of the set of joint angles is highly non-linear, and it typically presents hard constraints in the admissible joint angles.
Many of the data-driven methods used to learn the inverse kinematics learn a mapping $\fhat \colon \X \to \Y$ by minimizing some loss that measures the fidelity of the predicted $\hat{y}$ with respect to the training $\{y\}_{i=1}^{n}$ in the configuration space $\Y$. 
However, the goal of learning the inverse kinematics is often subordinated to a higher-level task aimed at controlling the robot in the Cartesian space $\X$.
Therefore, the true goal is to learn an inverse kinematics function $\fhat$ such that $g(\fhat(x)) \simeq x$, where $g$ is the true forward mapping of the robot.
To such end, we propose a novel structured prediction approach that faithfully encodes the structure of the configuration space (including constraints on the joints), but also evaluates the fidelity of the prediction in the Cartesian space. 
This leads to lower errors and more robust behaviour. 
To do so, we assume to have at our disposal the forward kinematics function of the robot at hand, even if with some errors, \textit{i.e.} $\gt \simeq g$.

Our approach provides a solution to problem~\cref{eq:ik} using the framework for structured prediction proposed in~\cite{ciliberto2017consistent}.
Given a set of joint configurations and corresponding end-effector poses $\mathcal{D} = ( x_i, y_i )_{i=1}^{n}$, generated by a forward kinematics function $g \colon \mathcal{Y} \to \mathcal{X}$, our goal is to \textit{learn} a function $\fhat(x) \approx g^{-1}$. 
Let $\loss \colon \mathcal{Y} \times \mathcal{Y} \to \mathbb{R}$ be a structured loss function that measures prediction errors in the output space. 
In the context of this work, $\Y$ corresponds to the space of (constrained) joint configurations. 
Let also $k:\X\times\X\to\R$ be a {\itshape kernel function} \cite{steinwart2008support} on input data (the target end-effector pose, in the setting of this work). 
We recall that kernel functions are a standard tool used in machine learning settings to learn non-linear non-parametric models (see \cite{steinwart2008support} and references therein). 
For context, several choices of kernel functions are typically available in practice, such as linear $k(x,x') = x^\top x'$, Gaussian $k(x,x') = e^{-\|x-x'\|^2/\sigma^2}$, or Laplacian $k(x,x') = e^{-\|x-x'\|/\sigma}$ (where $\sigma>0$ is a hyper-parameter). 
In all experiments reported in \cref{sec:experiments}, we employed a Gaussian kernel. This choice was motivated by the practical flexibility and theoretical properties (see \cref{sec:theory}) of such kernel. 

Then, the general form of a consistent regularized structured predictor (\crisp{}), as introduced in~\cite{ciliberto2017consistent}, is obtained by minimizing a data-dependent functional $F_{\alpha}$ of the form
\begin{equation}\label{eq:CR_estimator}
    \fhat(x) = \argmin_{y \in \Y}~ \left\{F_{\alpha(x)}(y) = \sum\limits_{i = 1}^{n} \alpha_i(x) \loss(y, y_i)\right\},
    % \fhat_{\crisp{}} (x) = \argmin_{y \in \Y} \sum\limits_{i = 1}^{n} \alpha_i(x) \loss(y, y_i),
\end{equation}
with weights $\alpha(x) = [\alpha_1, \ldots, \alpha_n]^\top = \big( \mathbf{K} + n\lambda \mathbf{I}_n \big)^{-1} \mathbf{K}_x$, where $\mathbf{K}\in\R^{n \times n}$ is the kernel matrix associated to the kernel $k$, with entries $\{\mathbf{K}\}_{ij} = k(x_i, x_j)$, and $\mathbf{K}_x \in \R^n$ is the evaluation vector with entries
$\{\mathbf{K}_x\}_i = k(x,x_i) $. 
Finally, $\mathbf{I}_n \in \R^{n \times n}$ is the identity matrix. In general, an estimator of the form in \cref{eq:CR_estimator} can be interpreted as mapping an input point $x$ to a $y = \fhat(x)$ corresponding to a {\itshape weighted barycenter} of the training output samples $y_i$ according to the loss $\loss$. The weights $\{\alpha_i\}_{i=1}^{n}$ define how relevant are the output examples $\{y_i\}_{i=1}^n$ for the considered test point.

\cref{alg:crisp} summarizes the two key steps characterizing CRiSP: training and prediction. First, during the training phase, we compute the inverse $A$ of the regularized kernel matrix $(\mathbf{K}+n\lambda \mathbf{I}_n)$. This process is akin to training a kernel ridge regression (KRR) model and can be carried out using any solver for linear systems (time complexity $\mathcal{O}(n^3)$ and memory complexity $\mathcal{O}(n^2)$). 
In our experiments, we compute the inverse of $A$ using its Cholesky decomposition to obtain a numerically robust solution.
%In our experiments we employed the FALKON package \cite{meanti2020kernel} \raf{@gianma is this right?}, which was recently developed to tackle KRR problems with high efficiency. 
% We note that kernel ridge regression, and therefore computationally similar algorithms, can be significantly accelerated by adopting recent randomized projection strategies without sacrificing statistical accuracy \cite{rudi2015less}. 
%While in this work we focus on the loss structure, we mention that t

Then, given a new input $x$, the prediction step in \cref{alg:crisp} consists in first computing the weights $(\alpha_i(x))_{i=1}^n$ via the matrix-vector product $\alpha(x) = A\mathbf{K}_x$ and then solving the constrained minimization problem in \cref{eq:CR_estimator}. In our experiments, we addressed this latter problem adopting the L-BFGS-B optimizer
%\footnote{\url{https://docs.scipy.org/doc/scipy/reference/generated/scipy.optimize.minimize.html}} 
\cite{byrd1995limited} from the \texttt{SciPy} scientific computing library~\cite{virtanen2020scipy}, which proved to be the most efficient among the constrained optimizers we tried in our experiments. For the purpose of reproducibility, we have made the implementation of CRiSP available to the community\footnote{ \url{https://github.com/gmmarconi/CRiSP-for-Misspecified-Robot-Model}}. 

The optimization of \cref{eq:CR_estimator} strongly depends on the loss $\loss$ employed. Below, we consider how to design such a loss in the context of learning the inverse kinematics of a robot.

\subsection{Choosing the Structured Loss $\loss$}

A first key question is how to choose the loss $\loss$ to measure prediction errors. In principle, it might be tempting to consider a loss such as the squared sum of joint angle differences, which naturally quantifies the difference between the joint configurations between the predicted and the measured joint configurations.
However, we argue that this might cause problems since configurations that are distant with respect to the metric on $\Y$ could correspond to similar poses in Cartesian space. 
This issue is illustrated by \cref{fig:loss_joints}, where the \crisp{} estimator trained with such configuration space loss is unable to predict a correct joint configuration to reach the desired target, and often shows a positional bias depending on the considered workspace region.

In contrast, here we propose a structured loss that measures how much two joint configurations ``differ'' in Cartesian space. 
More precisely, we assume that a --~possibly inaccurate~-- forward kinematics function $\gt(y) = [\gt_p(y), \gt_o(y)]^\top$ is available, where $\gt_p \colon \mathcal{Y} \to  \mathbb{R}^d$ and $\gt_o \colon \mathcal{Y} \to SO(d)$ are the components that map respectively to the position and the orientation of the end effector.
It is important to note that this function can be different from the ground-truth forward kinematics $g$ used to generate the dataset $\mathcal{D}$.
Then, we propose the {\itshape forward kinematics loss (FK)}
\begin{equation}\label{eq:fk_loss}
\loss(y, y_i) = \| \gt_p(y) - \gt_p(y_i) \|^2 + d_{O} (\gt_o(y), \gt_o(y_i))^2,
\end{equation}
where the first term measures the Euclidean distance of the end effector from the desired position, while the second term
\begin{equation}\label{eq:circle_distance}
d_{O}(y, z)^2 := \sum\limits_{j=1}^{c} \min(|y_j - z_j|, 2\pi-|y_j - z_j|)^2
\end{equation}
measures the error between the predicted and the target end-effector orientation with respect to the squared geodesic distance on the circle, which can be used as an alternative representation for $SO(d)$, with $c = 1$ for $SO(1)$ and $c=3$ for $SO(3)$.
Note that it is also possible to weight differently the elements in~\cref{eq:fk_loss} to adjust position and orientation accuracies according to the desired performance.
%, but we did not investigate this effect in our experiments. \raf{beh qualche effetto l'abbiamo visto.}

We refer to \crisp{}-FK as the estimator employing such loss. 
\cref{fig:loss_cartesian} shows that, as designed, this estimator is better-suited to learn the inverse kinematics of a robotic manipulator.

\begin{algorithm}[t]
\caption{CRiSP\label{alg:crisp}}
\begin{algorithmic}
\STATE \hspace{-1em}{\bfseries Input:} $\mathcal{D}= (x_i,y_i)_{i=1}^n$ training set, $k$ kernel, $\lambda>0$ regularization parameter. 

\vspace{1em}

\STATE \hspace{-1em}{\bfseries Training:}
\STATE Compute the kernel matrix $(\mathbf{K})_{ij} = k(x_i,x_j)$
\STATE Compute $A = (\mathbf{K} + n\lambda \mathbf{I}_n)^{-1} \in\R^{n \times n}$ 
%// e.g. \cite{meanti2020kernel}

\vspace{1em}
\STATE \hspace{-1em}{\bfseries Prediction.} For any new input $x$:
\STATE Evaluate $\mathbf{K}_x = (k(x,x_1),\dots,k(x,x_n))^\top \in \R^n$
\STATE Compute the weights $\alpha(x) = A \mathbf{K}_x$.
\STATE $\hat y =$\texttt{ Minimize}$(F_{\alpha(x)}(\cdot))$\quad// e.g. use L-BFGS~\cite{byrd1995limited}
\STATE {\bfseries Return:} $\hat y$.
\end{algorithmic}
\end{algorithm}

\subsection{Statistical Properties of \crisp{}} \label{sec:theory}
By leveraging the structured prediction perspective from \cite{ciliberto2020general}, here we characterize the statistical properties of the proposed \crisp{} estimator. In particular, the following result proves that $\fhat$ is {\itshape universally consistent}, namely that as the number $n$ of training examples grows, $\fhat$ is guaranteed to asymptotically converge to the ideal inverse kinematics $f_*:\X\to\Y$ minimizing
\eqal{\label{eq:fstar-solution}
f_* = \argmin_{f:\X\to\Y}~ \mathbb{E}_\rho ~\loss(f(x),y),
}
where $\rho$ is the probability distribution on $\X\times\Y$ from which we sample the train-test pairs $(x,y)$. More formally, we have the following result.

\begin{theoremdecente}[Universal Consistency]\label{thm:consistency} Let $\mathcal{D} = (x_i, y_i)_{i=1}^{n}$ be sampled i.i.d. according to a distribution $\rho$ on $\X\times\Y$, let $\fhat$ be the estimator in \cref{eq:CR_estimator} trained with FK loss $\loss$ from \cref{eq:fk_loss} and $\lambda = n^{-1/2}$ on $\mathcal{D}$, using a universal kernel \cite{steinwart2008support} (e.g. Gaussian or Laplacian). Then, let $f_*$ the ideal inverse kinematics from \cref{eq:fstar-solution}, then with probability $1$,
\eqal{
    \lim_{n\to+\infty}~\mathbb{E}_\rho~ \loss(\fhat(x),y) ~=  \mathbb{E}_\rho~ \loss(f_*(x),y).
}
\end{theoremdecente}

\begin{proofdecente}
The result requires showing that the $\loss$ satisfies the {\itshape Implicit Loss Embedding (ILE)} property \cite[Def. 1]{ciliberto2020general} (a technical property whose definition is outside the scope of this paper. We refer the interested reader to the original work). We first note that such property holds already for the squared difference $\|\cdot-\cdot\|^2$ and orientation loss $d_O(\cdot,\cdot)^2$ satisfy such property (see e.g. \cite{rudi2018manifold}). 
It follows that also the FK loss $\loss$ satisfies the ILE property, since sum and composition with smooth functions (namely $\gt_p$ and $\gt_o$) still satisfy the ILE property \cite[respectively Thm. 10 and Cor. 11]{ciliberto2020general}. Then, \cref{thm:consistency} follows as a direct corollary of \cite[Thm. 4]{ciliberto2020general}.
\end{proofdecente}

\noindent\cref{thm:consistency} shows that the proposed algorithm asymptotically yields the \textit{best} possible inverse kinematics map $f_*$ {\itshape with respect to the training distribution $\rho$}. By introducing additional regularity assumptions on the inverse kinematics map, it is possible to improve the result above, yielding also {\itshape non-asymptotic rates}. In particular, we require that the ideal inverse kinematics $f_*$ belongs to a suitable Sobolev space $W^{s,2}$ of square-integrable functions with the first $s\in\R$ weak derivatives square-integrable (see \cite{adams2003sobolev} for a formal definition). 
The latter is a standard assumption in supervised learning settings \cite{ciliberto2016consistent}, and essentially requires the inverse kinematics to be a regular function (e.g. smooth with controlled derivatives). 
Then, the following result concludes our analysis, providing upper bounds on \crisp{}'s excess risk.

\begin{theoremdecente}[Rates]\label{thm:rates}
With the same hypotheses of \cref{thm:consistency} and using a Laplacian kernel, assume that the ideal solution $f_*\in W^{s,2}(\R^d)$, with $s>d/2$. Then, with high probability with respect to $\rho$
\eqal{\label{eq:rates}
    \mathbb{E}_\rho~ [\loss(\fhat(x),y) - \loss(f_*(x),y)] \leq O(n^{-1/4}).
}
\end{theoremdecente}
\begin{proofdecente}
The proof is analogous to that of \cref{thm:consistency} but relies on the assumption $f^*\in W^{s,2}(\R^d)$ to apply \cite[Thm. 5]{ciliberto2020general} to the \crisp{} estimator, yielding \cref{eq:rates} as required. 
\end{proofdecente}

%===============================================================================

\section{Experiments}\label{sec:experiments}

In this section, we empirically validate the proposed approach.
In particular, we show that:
\begin{itemize}
\item The flexibility provided by Structured Prediction allows \crisp{}-FK to overcome the limitations of other data-driven methods.
In particular, in Sec.~\ref{subsec:fk-vs-joints-loss} we show that being able to define a structured loss directly in Cartesian space allows our method to outperform other Machine-Learning-based IK approaches that employ losses in joint configuration space.
\crisp{}-FK also yields IK solutions respecting joint position limits by construction, by simply defining box constraints in \eqref{eq:CR_estimator}.
\item Thanks to the previously introduced properties, \crisp{}-FK is more robust to model misspecification with respect to model-based IK and consistently outperforms it in several 2D and 3D settings, as reported in Sec.~\ref{subsec:misspecified-model-exp}.
\end{itemize}

\begin{figure*}[!t]
    \centering
    \begin{subfigure}[t]{.24\linewidth}
        \centering
    	\includegraphics[width=\textwidth]{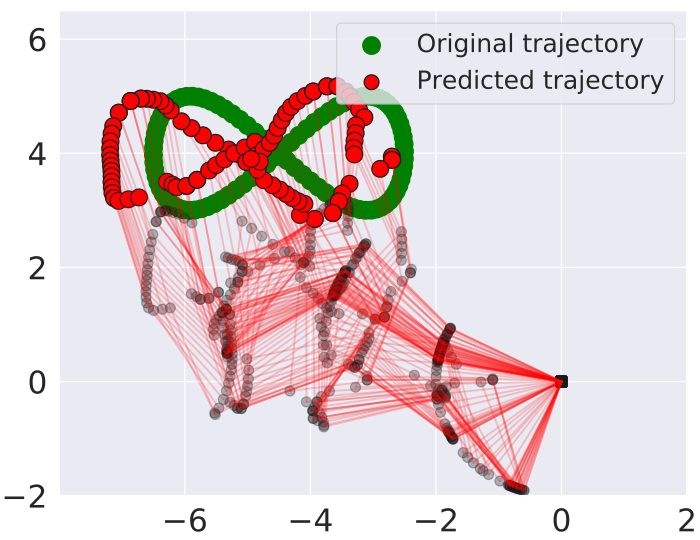}
    	\caption{OCSVM}
        \label{fig:compare_OCSVM}
    \end{subfigure}
    ~
    \hfill
        \begin{subfigure}[t]{.22\linewidth}
        \centering
        \includegraphics[width=\textwidth]{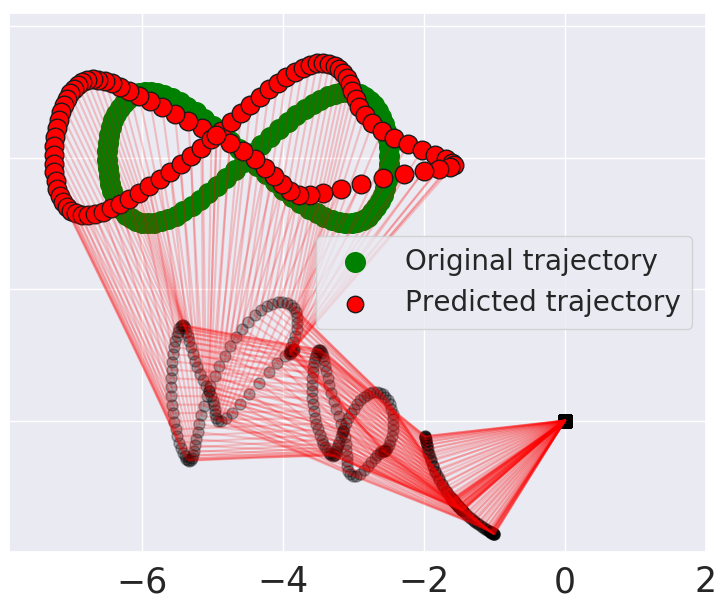}
        \caption{Neural Network}
        \label{fig:compare_nn}
    \end{subfigure}
    ~
    \hfill
    \centering
    \begin{subfigure}[t]{.22\linewidth}
        \centering
    	\includegraphics[width=\textwidth]{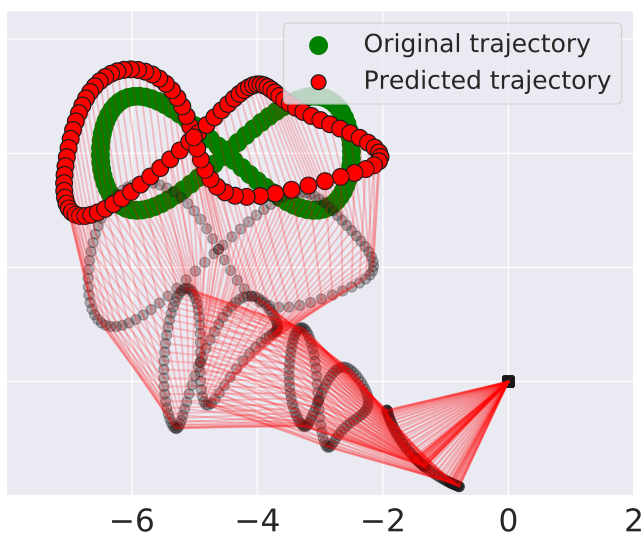}
    	\caption{\crisp{}-R}
        \label{fig:compare_crisp}
    \end{subfigure}
    ~
    \hfill
    \begin{subfigure}[t]{.22\linewidth}
        \centering
        \includegraphics[width=\textwidth]{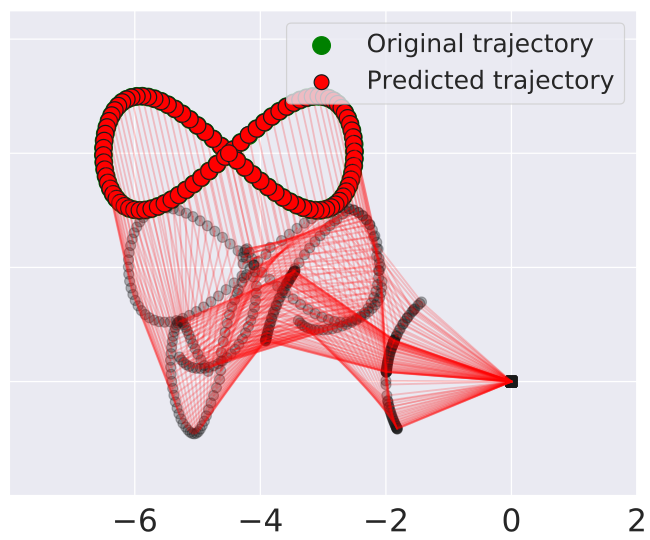}
        \caption{\crisp{}-FK}
        \label{fig:compare_crisp_fk}
    \end{subfigure}
    \caption{\small Qualitative results for trajectory reconstruction on a planar robotic manipulator of One Class Structured SVM {\bfseries (a)}, the neural network model{\bfseries (b)},  \crisp{} with radians loss on the joints {\bfseries (c)} and \crisp{} with the forwards kinematics loss {\bfseries (d)}. Scale in meters.}
    \label{fig:planar_fk-vs-joints}
\end{figure*}

\subsection{Target Task: Trajectory Reconstruction}

We evaluate the performance of our approach on the trajectory reconstruction task described below. 
By using an estimator in the form of~\cref{eq:struct_generic_estimator}, it is possible to instantiate a single global model over the whole Cartesian space, where each local minimum of $F_\alpha(x,y)$ corresponds to a possible solution. 
This overcomes the problem of the non-injectivity of $g$, since there is always a unique local solution found by gradient-based iterative optimizers such as L-BFGS-B, depending on the starting $y_0$. 
This property can be leveraged in common robotics tasks such as trajectory reconstruction, where it is necessary to solve the inverse kinematics for a sequence of points $\{x_t\}_{t=1}^{L}$ which describes a trajectory in space. 
We follow the idea from~\cite{bocsi2011learning} and compute the inverse kinematics of a trajectory one point at a time, using the inferred $\fhat(x_{t-1})$ of the previous trajectory point as the starting  L-BFGS-B value for the prediction of the next point in the trajectory.
%\raf{@gianma: alla fine hai fatto girare gli esperimenti definitivi con l'hot restart?}
This idea hinges on the assumption that by starting from a configuration $y_t$, the solution $y_{t+1}$ for the next point will be similar, providing continuity in the joints movement.

\subsection{Experimental Setup}\label{sec:robot-setup}
In our first set of experiments we employ a 5-DoF simulated planar manipulator with $5$ links of $2~m$ each, and $5$ revolute joints spanning the output space (in radians) $\Y_5 = [0, \pi] \times  [-\pi, 0] \times [-\pi/2, -\pi/2] \times [-\pi, 0] \times [-\pi/2, \pi/2]$. 
We generate training and validation sets by sampling $n = 25000$ random configurations $\{y_i\}_{i=1}^n$ with uniform distribution in $\Y_5$ and compute the corresponding poses $\{g_5(y_i)\}_{i=1}^n = \{x_i\}_{i=1}^n$, with $x_i \in \R^2 \times SO(1) \quad \forall i = 1,\ldots,n$. 
Since $g_5$ is a highly non-linear function, while the sampled configurations have uniform distribution, the corresponding poses are not uniformly distributed.
We aim at reconstructing an eight-shaped trajectory, well within the robot's workspace, and a circle-shaped trajectory, closer to the boundary of the reachable region.

The second set of experiments was executed in 3D Cartesian space, and it involves realistic IK tasks.
To this aim, we use a simulated 7-DoFs Franka Emika Panda manipulator (shown in Fig.~\ref{fig:panda-frontpage}).
We employed the official joint position limits throughout our experiments, yielding the following configuration space $\Y = [-2.90, 2.90] \times  [-1.76, 1.76] \times [-2.90, -2.90] \times [-3.07, -0.07] \times [-2.90,  2.90] \times [-0.02, 3.75] \times [-2.90, -2.90]$. Note that all quantities are expressed in radians and rounded to the third digit. We consider two datasets and two test trajectories for assessing the performances of our approach.
For the first task, we uniformly sample 35000 training points from a solid torus with radius 1~$cm$ around a circular trajectory centered in [0.0, 4.4, -55]~$cm$ and with radius 3~$cm$.
Each sample has either one of two fixed orientations with respect to the $z$ axis of the world reference frame: $+\pi/4$~$rad$ and $-\pi/4$~$rad$.

The second task involves the reconstruction of a spiraling trajectory of radius 3~$cm$ and height 6~$cm$.
We uniformly sample 35000 training points, but this time from a larger region of the workspace so as to have a lower density of examples per unit of volume.
Moreover, the collected end-effector orientations are no longer constrained, making the learning task more challenging and high-dimensional.
We have experimentally observed  that our method relies on local information from training samples, which suggests that restricting our experiments to specific areas of Cartesian space does not imply a loss in generality as long as the robot is not too close to singular configurations.
However, having a lower sample density sets-up a harder problem.

% $\Y_7 = [-2.8973, 2.8973] \times  [-1.7628, 1.7628,] \times [-2.8973, -2.8973] \times [-3.0718, -0.0698] \times [-2.8973,  2.8973] \times [-0.0175, 3.7525] \times [-2.8973, -2.8973]$.
Our experiments are based on the following software components.
We use the Bullet 3  simulator~\cite{coumans2016pybullet} (via the PyBullet Python interface) for simulating our robots and tasks.
We employ the Selectively-Damped Least Squares (SDLS) algorithm~\cite{buss2005selectively} available in Bullet as an IK baseline for our experiments.
Hyperparameter selection for \crisp{}-FK was performed via grid search on a separate validation set.
% We released our implementation of CRiSP, together with the experimental datasets, at \url{https://github.com/gmmarconi/CRiSP-for-Misspecified-Robot-Model}.

\begin{table}
    \centering
    \caption{
    % \scriptsize 
    Root mean square error for position (in cm) and orientation (radians) on two test trajectories: eight and circle. Results reported are for the neural network model (NN), One Class Structured SVM (OCSSVM), \crisp{}-R, and \crisp{}-FK}
    % \resizebox{\linewidth}{!}{%
            \begin{tabular}{c|c|c|c|c|c}
            % \toprule
            Traj. &	& OCSVM & NN & \crisp{}-R  & \crisp{}-FK\\
            \midrule
            % \cmidrule(r){2-6}
            \multirow{2}{*}{Eight} 
            & Pos. [cm] & $57\pm15$ & $88\pm24$ & $80\pm14$ & \textbf{0.05 ± 0.04} \\
            & Orn. [rad] & $0.18\pm0.38$ & $0.04\pm0.01$ & $0.07\pm0.07$ & \textbf{0.03 ± 0.03}  \\
            % \cmidrule(r){2-5}
            \midrule
            \multirow{2}{*}{Circle}
            & Pos.  [cm] & $275\pm95$  & $81\pm10$ & $54\pm14$ & \textbf{0.04 ± 0.01} \\
            & Orn.  [rad] & $0.08\pm0.06$ &  $0.05+0.01$ &\textbf{0.02±0.02} & $0.03\pm 0.01$ 
            % \bottomrule
            \end{tabular}
    % }
    \vspace{-0.2cm}
    \label{tab:FK-vs-joints_losses}
\end{table}

\subsection{Comparison of Data-driven Methods}
% \subsection{Forward Kinematics Loss Versus Joint-space Loss}
\label{subsec:fk-vs-joints-loss}

We assess the impact of the two different choices of loss function from both a qualitative and quantitative perspective. For reference, we also compare \crisp{} with a neural network estimator that does not take into account the structure of the robot (i.e. the joint constraints) as well as with the One-Class Structured SVM (OCSSVM) introduced in \cite{bocsi2011learning}. For the neural model, we trained a five-layer neural network (NN) comprised of fully connected layers and hyperbolic tangent activations, where the first two and last two layers have 64 hidden units each, while the middle layers have 128 hidden units each. The NN performs multivariate regression directly into the joints space and does not take into account the joint constraints of the robot. When the NN produces joint configurations outside the robot's constraint, we clip the predictions to satisfy the constraints.  Regarding our proposed estimator we consider: 1) a \crisp{} predictor with loss $d_O$ corresponding to the sum of squared radial distances between joint angles from~\cref{eq:circle_distance};
2)~the \crisp{} model with FK loss $\loss$ introduced in \cref{eq:fk_loss}. We refer to these models as \crisp{}-R and \crisp{}-FK, respectively.We cross-validate for hyperparameters on $10000$ randomly-sampled  validation points  and evaluate the performance on two test trajectories: an eight-shaped one and a circular one.

\cref{fig:planar_fk-vs-joints} offers a qualitative comparison between the different methods, showing the corresponding predictions in the case of tracking the eight-shaped trajectory. 
It can be noticed that all methods but \crisp{}-FK are unable to correctly track the required trajectory. This is likely due to the fact that \crisp{}-FK combines the best of both data-driven and model-based approaches, by employing a loss function comparing the resulting Cartesian poses instead of the joint values. These observations are quantitatively supported in \cref{tab:FK-vs-joints_losses}, which reports the average prediction error in both position and orientation of the different methods.  

We note that OCSSVM performs particularly poorly on the circular trajectory. We argue that this is due to intrinsic limitations of OCSSVM, which was originally developed for support estimation purposes and therefore needs a much higher sample complexity. In fact, the circular trajectory is very close to the boundaries of the manipulator workspace, where training samples are sparse.
Given the empirical observations in this section, in the following we do not report additional results for the NN model, OCSSVM, and \crisp{}-R,  since their performances are sub-optimal with respect to the proposed \crisp{}-FK strategy (and in the case of OCSSVM, extremely demanding in terms of computational time). Rather, we focus on a comparison with model-based methods, which are more computationally efficient.

\begin{figure}[t]
    \centering
    \begin{subfigure}[t]{.48\linewidth}
        \centering
        \includegraphics[width=\textwidth]{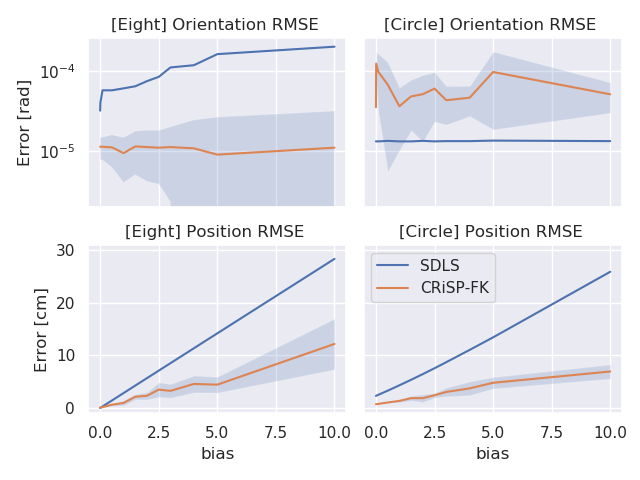}
        \subcaption{Biased links}
        \label{fig:planar_biased_length_plots}
    \end{subfigure}
    ~
    \begin{subfigure}[t]{.48\linewidth}
        \centering
        \includegraphics[width=\textwidth]{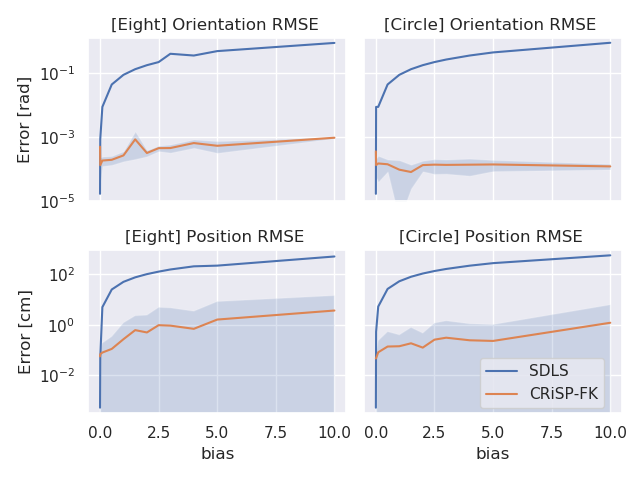}
        \subcaption{Biased joints}
        \label{fig:planar_biased_plots}
  \end{subfigure}
  \caption{On the $y$ axis the errors for orientation (in radians) and position (in cm) of circle-shaped and eight-shaped trajectories for models with bias in the links (\cref{fig:planar_biased_length_plots}) and bias in the joints (\cref{fig:planar_biased_length_plots}. On the $x$ axis the magnitude of the bias in cm for the links and degrees for the joints.}
\end{figure}

% \usepackage{graphicx}
\begin{table*}[]
\centering
\caption{\small
IK performances comparison of \crisp{}-FK and SDLS on 4 desired trajectories (position and orientation) for the end-effector (EE) of a 7-DoF Panda arm.
$S$: spiral with fixed desired EE orientation;
% , with CRiSP trained on data sampled from a cube with unconstrained EE orientations
$C_0$, $C_{-\pi/4}$, $C_{+\pi/4}$: Circumference with 3 different EE orientations.
% , trained on data sampled from a solid torus surrounding the circumference.
The Panda arm's kinematic model is corrupted by introducing increasing biases at each joint (0 to 3~$deg$) or at each link (0 to 30~$mm$).
Position ($mm$) and orientation ($rad \cdot 10^{-4}$) root mean square errors (RMSE) are reported with one standard deviation along the trajectories.}
\resizebox{\textwidth}{!}{%
\begin{tabular}{cc|ll|ll|ll|ll}
\multicolumn{1}{r}{\textit{}}            & \textit{Trajectory}                                 & \multicolumn{2}{c|}{\textbf{$S$}}                                       & \multicolumn{2}{c|}{\textbf{$C_0$}}                                     & \multicolumn{2}{c|}{\textbf{$C_{-\pi/4}$}}                              & \multicolumn{2}{c}{\textbf{$C_{+\pi/4}$}}\\
\cline{2-10} 
\multicolumn{1}{r}{\textit{}}            & \textit{Algorithm}                                   & \multicolumn{1}{c}{\textbf{CRiSP-FK}} & \multicolumn{1}{c|}{\textbf{SDLS}} & \multicolumn{1}{c}{\textbf{CRiSP-FK}} & \multicolumn{1}{c|}{\textbf{SDLS}} & \multicolumn{1}{c}{\textbf{CRiSP-FK}} & \multicolumn{1}{c|}{\textbf{SDLS}} & \multicolumn{1}{c}{\textbf{CRiSP-FK}} & \multicolumn{1}{c}{\textbf{SDLS}} \\ \hline
\multicolumn{1}{c|}{\textit{Joint Bias (deg)}} & \textit{}                                      & \multicolumn{1}{c}{\textit{}}      &                                    &                                    &                                    &                                    &                                    &                                    &                                   \\
\multicolumn{1}{c|}{0}                   & \textit{}                                            & \textbf{0.739 ± 0.303}             & 1.33 ± 0.386                       & \textbf{0.602 ± 0.173}             & 0.749 ± 0.142                      & \textbf{0.548 ± 0.153}             & 0.762 ± 0.262                      & \textbf{0.576 ± 0.124}             & 3.58 ± 0.385                      \\
\multicolumn{1}{c|}{0.01}                & \textit{Pos. RMSE}                                   & \textbf{0.749 ± 0.309}             & 1.42 ± 0.393                       & \textbf{0.623 ± 0.173}             & 1.05 ± 0.141                       & \textbf{0.56 ± 0.161}              & 0.61 ± 0.208                       & \textbf{0.575 ± 0.126}             & 3.67 ± 0.388                      \\
%\multicolumn{1}{c|}{0.05}                & \textit{}                                           & \textbf{0.92 ± 0.431}              & 1.81 ± 0.392                       & \textbf{0.765 ± 0.126}             & 2.28 ± 0.139                       & \textbf{0.588 ± 0.172}             & 1.3 ± 0.03                         & \textbf{0.574 ± 0.125}             & 4.41 ± 0.154                      \\
\multicolumn{1}{c|}{0.1}                 & \textit{($mm$)}                                      & \textbf{1.17 ± 0.641}              & 2.99 ± 0.294                       & \textbf{1.01 ± 0.155}              & 3.81 ± 0.141                       & \textbf{0.623 ± 0.181}             & 2.93 ± 0.05                        & \textbf{0.573 ± 0.126}             & 5.75 ± 0.165                      \\
%\multicolumn{1}{c|}{0.5}                 & \textit{(mm)}                                       & \textbf{3.75 ± 2.74}               & 14 ± 0.481                         & \textbf{3.6 ± 0.515}               & 16.1 ± 0.269                       & \textbf{1.45 ± 0.819}              & 15.9 ± 0.216                       & \textbf{0.568 ± 0.13}              & 17.1 ± 0.344                      \\
\multicolumn{1}{c|}{1}                   & \textit{}                                            & \textbf{7.31 ± 5.37}               & 27.9 ± 0.824                       & \textbf{7.26 ± 1.25}               & 31.7 ± 0.5                         & \textbf{1.8 ± 1.06}                & 32.1 ± 0.421                       & \textbf{0.56 ± 0.137}              & 35.2 ± 0.563                      \\
%\multicolumn{1}{c|}{2}                   & \textit{}                                           & \textbf{14.3 ± 10.5}               & 55.7 ± 1.61                        & \textbf{13.7 ± 2.21}               & 63.3 ± 0.959                       & \textbf{3.85 ± 2.8}                & 64.6 ± 0.818                       & \textbf{0.57 ± 0.116}              & 72.3 ± 1.03                       \\
\multicolumn{1}{c|}{3}                   & \textit{}                                            & \textbf{21.2 ± 15.5}               & 83.7 ± 2.48                        & \textbf{20.1 ± 3.44}               & 95.4 ± 1.38                        & \textbf{9.78 ± 7.95}               & 97 ± 1.2                           & \textbf{0.592 ± 0.106}             & 110 ± 1.47                        \\ \hline
\multicolumn{1}{c|}{\textit{Joint Bias (deg)}} & \multicolumn{1}{l|}{}                          &                                    &                                    &                                    &                                    &                                    &                                    &                                    &                                   \\
\multicolumn{1}{c|}{0}                   & \textit{}                                            & 32.2 ± 19.3                        & \textbf{7.28 ± 2.29}               & 9.4 ± 0.77                         & \textbf{3.92 ± 0.95}               & \textbf{2.92 ± 0.71}               & 6.97 ± 0.11                        & \textbf{19.9 ± 1.35}               & 27.9 ± 1.03                       \\
\multicolumn{1}{c|}{0.01}                & \textit{Orn. RMSE}                                   & 32.2 ± 19.3                        & \textbf{7.46 ± 2.19}               & 9.61 ± 0.66                        & \textbf{4.08 ± 0.48}               & \textbf{2.99 ± 0.75}               & 10 ± 0.21                          & \textbf{19.9 ± 1.35}               & 29.2 ± 1.17                       \\
%\multicolumn{1}{c|}{0.05}                & \multicolumn{1}{l|}{}                               & 35.1 ± 19.8                        & \textbf{20.9 ± 0.51}               & \textbf{12.8 ± 1.1}                & 22.4 ± 0.18                        & \textbf{3.46 ± 0.65}               & 26.9 ± 0.93                        & \textbf{19.9 ± 1.39}               & 34.6 ± 1.79                       \\
\multicolumn{1}{c|}{0.1}                 & \textit{($rad \cdot 10^{-4}$)}                                  & 40.4 ± 21.3                        & \textbf{41.2 ± 1.06}               & \textbf{20 ± 2.52}                 & 45.6 ± 0.33                        & \textbf{4.4 ± 1.45}                & 49.6 ± 1.76                        & \textbf{19.9 ± 1.41}               & 54.3 ± 1.16                       \\
%\multicolumn{1}{c|}{0.5}                 & \multicolumn{1}{l|}{\textit{(rad $\cdot 10^{-4}$)}} & \textbf{109 ± 48.3}                & 203 ± 5.8                          & \textbf{95.6 ± 8.56}               & 231 ± 1.46                         & \textbf{25.1 ± 19.3}               & 230 ± 8.4                          & \textbf{20.3 ± 1.51}               & 245 ± 2.85                        \\
\multicolumn{1}{c|}{1}                   & \multicolumn{1}{l|}{}                                & \textbf{203 ± 86.6}                & 403 ± 11.9                         & \textbf{198 ± 22.8}                & 462 ± 2.72                         & \textbf{32.8 ± 24.3}               & 453 ± 16.7                         & \textbf{21.1 ± 1.81}               & 485 ± 4.38                        \\
%\multicolumn{1}{c|}{2}                   & \multicolumn{1}{l|}{}                               & \textbf{390 ± 165}                 & 797 ± 24.7                         & \textbf{377 ± 38.9}                & 919 ± 4.84                         & \textbf{73.7 ± 57.7}               & 885 ± 33.1                         & \textbf{23.1 ± 3.15}               & 974 ± 7.82                        \\
\multicolumn{1}{c|}{3}                   & \multicolumn{1}{l|}{}                                & \textbf{611 ± 260}                 & 1.18e+03 ± 38.2                    & \textbf{548 ± 59}                  & 1.37e+03 ± 6.61                    & \textbf{189 ± 157}                 & 1.3e+03 ± 49.5                     & \textbf{25 ± 4.89}                 & 1.47e+03 ± 11.4               \\ \hline
\multicolumn{1}{c|}{\textit{Link Bias (mm)}} & \textit{}                                        & \multicolumn{1}{c}{\textit{}}      &                                    &                                    &                                    &                                    &                                    &                                    &                                   \\
\multicolumn{1}{c|}{0.1}                & \textit{}                                             & \textbf{0.943 ± 0.248}              & 1.19 ± 0.246                       & \textbf{0.807 ± 0.123}             & 1.4 ± 0.055                        & \textbf{0.56 ± 0.161}              & 1.202 ± 0.036                      & \textbf{0.574 ± 0.121}             & 4.27 ± 0.390                      \\
\multicolumn{1}{c|}{1}                 & \textit{Pos. RMSE}                                     & \textbf{3.28 ± 0.988}               & 11.8 ± 0.204                       & \textbf{4.00 ± 0.097}              & 12.1 ± 0.086                       & \textbf{0.639 ± 0.174}             & 8.86 ± 0.026                       & \textbf{0.567 ± 0.128}             & 10.4 ± 0.463                      \\
\multicolumn{1}{c|}{10}                   & \textit{($mm$)}                                     & \textbf{30.5 ± 10.2}                & 121 ± 1.26                         & \textbf{37.8 ± 0.737}              & 120 ± 0.876                        & \textbf{2.89 ± 1.95}               & 90.0 ± 0.639                       & \textbf{0.518 ± 0.111}             & 93.1 ± 1.54                      \\
\multicolumn{1}{c|}{30}                   & \textit{}                                           & \textbf{88.3 ± 29.6}                & 330 ± 6.56                         & \textbf{115 ± 2.56}                & 349 ± 3.37                         & \textbf{2.96 ± 0.946}              & 286 ± 3.65                         & \textbf{0.586 ± 0.158}             & 296 ± 2.81                        \\ \hline
\multicolumn{1}{c|}{\textit{Link Bias (mm)}} & \multicolumn{1}{l|}{}                               &                                    &                                    &                                    &                                    &                                    &                                    &                                    &                                   \\
\multicolumn{1}{c|}{0.1}                & \multicolumn{1}{l|}{}                                 & 24.3 ± 11.3                         & \textbf{7.30 ± 2.29}               & 9.39 ± 0.77                        & \textbf{3.93 ± 0.95}               & \textbf{2.94 ± 0.79}               & 6.98 ± 0.11                        & \textbf{19.9 ± 1.35}               & 27.9 ± 1.03                       \\
\multicolumn{1}{c|}{1}                 & \textit{Orn. RMSE}                                     & 24.4 ± 11.4                         & \textbf{7.52 ± 2.33}               & 9.44 ± 0.76                        & \textbf{4.03 ± 0.96}               & \textbf{2.92 ± 0.77}               & 7.03 ± 0.11                        & \textbf{19.9 ± 1.32}               & 27.7 ± 1.04                       \\
\multicolumn{1}{c|}{10}                   & \textit{($rad \cdot 10^{-4}$)}                      & 24.4 ± 11.3                         & \textbf{9.94 ± 2.31}               & 9.37 ± 0.80                        & \textbf{5.21 ± 1.05}               & \textbf{3.04 ± 0.72}               & 7.43 ± 0.16                        & \textbf{19.9 ± 1.32}               & 25.9 ± 1.10                        \\
\multicolumn{1}{c|}{30}                   & \multicolumn{1}{l|}{}                               & 24.3 ± 11.4                         & \textbf{11.0 ± 0.54}               & \textbf{9.77 ± 1.00}               & \textbf{8.25 ± 0.72}               & \textbf{3.49 ± 0.95}               & 7.37 ± 0.40                        & \textbf{20.0 ± 1.35}               & 21.8 ± 1.13               
\end{tabular}%
}
\label{tab:panda_exp}
\end{table*}

\subsection{Misspecified Model Compensation}\label{subsec:misspecified-model-exp}
\begin{figure}[]
    \begin{center}
    \begin{subfigure}[]{.4\textwidth}
        \centering
    	\includegraphics[width=\textwidth]{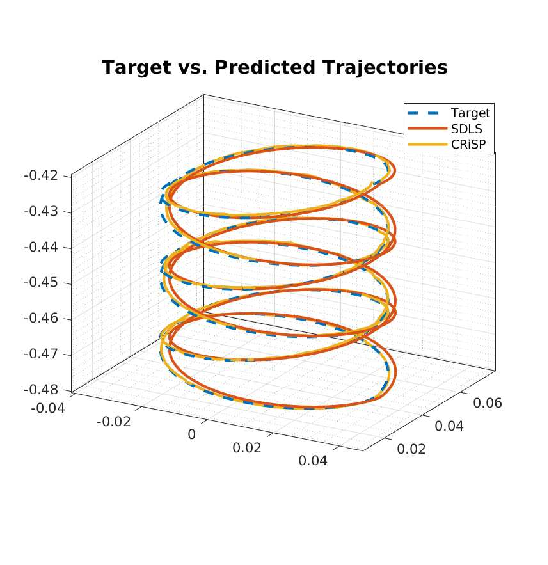}
    	\caption{}
        \label{fig:spiral_tracking}
    \end{subfigure}
    ~\qquad\quad
    % \hspace{-3em}
    \begin{subfigure}[]{.3\textwidth}
        \centering
        \includegraphics[width=\textwidth]{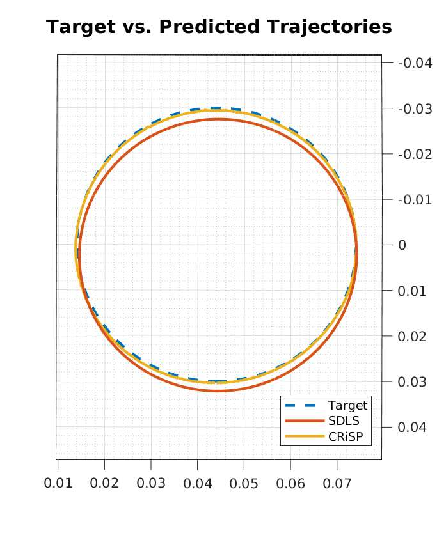}
        \caption{}
        \label{fig:circ_tracking}
    \end{subfigure}
    \caption{\small
     Trajectories reconstructed by CRiSP-FK and SDLS under model bias (0.1 $deg$) for the $S$ (a) and $C_0$ (b) tasks, respectively.}
  \label{fig:pandatraj}
  \end{center}
\end{figure}

\begin{figure}[t]
    \centering
     \begin{subfigure}[]{.45\textwidth}
        \centering
    	\includegraphics[width=\textwidth]{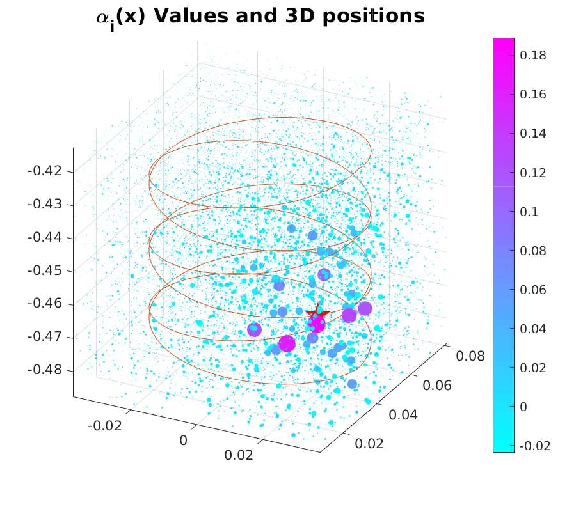}
    	\caption{}
        \label{fig:spiral_alphas}
    \end{subfigure}
    ~
    %\qquad\qquad\qquad
    \begin{subfigure}[]{.5\textwidth}
        \centering
        \includegraphics[width=\textwidth]{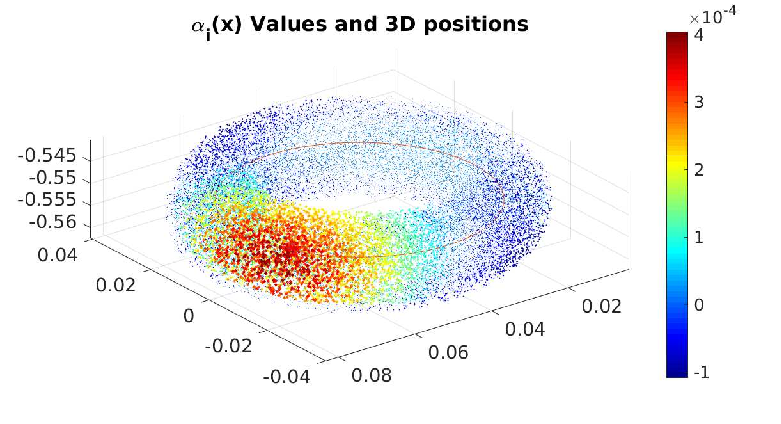}
        \caption{}
        \label{fig:circ_alphas}
    \end{subfigure}
    \caption{\small
    (a) and (b) show \crisp{}-FK's $\alpha_i(x)$ values for a representative spiral and circular trajectory point, respectively.}
\end{figure}

We evaluate the capability of \crisp{}-FK to compensate for errors in the supplied forward kinematics model $g$. To do so, we generate a dataset $\D$ using the true forward model $g$ of the robot and then we train our model using the loss $\loss(y, y_i) = \| \gt_p(y) - \gt_p(y_i) \|^2 + d_{O} (\gt_o(y), \gt_o(y_i)))^2$, where $\gt$ is computed as the original $g$, plus a fixed bias either in the joint angles or in the link lengths. 
For joint angles, if a configuration has values $y = [y_1,\ldots, y_J]$, then $\gt(y) = g(y + \bar{b})$ where $\bar{b} = b \cdot [1, \ldots, 1] \in \R^J $ and $b \in \R$ controls the amount of bias. 
For link lengths, we add $b$ to the nominal link lengths used to build the true $g$ , with the signs of $b$ chosen at random at every experiment repetition.
We test the trained model on trajectory reconstruction and report qualitative and quantitative results for increasing values of $b$. 
As a comparison baseline, we approach the same task using the SDLS inverse kinematics solver of PyBullet. 
This experiment is performed with both the planar manipulator and the Panda robot, training on the datasets and trajectories described in~\cref{sec:robot-setup}.
\cref{fig:planar_biased_plots} and \cref{fig:planar_biased_length_plots} show the RMSE error in position and orientation on both trajectories as a function of, respectively, the bias in the joints and in the links.
\cref{tab:panda_exp} shows the same type of error on four different trajectories: the spiral ($S$) and the circular trajectory with three different end-effector orientations ($C_{\pi/4}, C_{-\pi/4}, C_{0}$), where $C_{0}$ is an out-of-sample orientation that does not appear in the training set. In~\cref{fig:circ_alphas} and~\cref{fig:spiral_alphas} we show the magnitudes of $\alpha(x)$ from~\cref{eq:CR_estimator} to provide a visual intuition on the datasets and the relevance of each training sample during prediction. 
Finally, in~\cref{fig:circ_tracking} and~\cref{fig:spiral_tracking} we show an example of reconstructed trajectory with misspecified model.

\subsection{Results discussion} \label{sec:discussion}

The performance of CRiSP-FK is validated by the results for $0$ bias in~\cref{tab:panda_exp}, where the presented problem has more degrees of freedom than the planar one, and is, therefore, more challenging. 
In the case of the circular trajectory,~\cref{fig:circ_alphas} shows that even when the local sample density is high, only close examples are actually used to compute the proposed solution. 
When presented with a task for which the dataset is sparser, such as the spiral trajectory shown in~\cref{fig:spiral_tracking}, the performance is worse than SDLS for orientation but still better in position. 
A qualitative overview of our experiments is reported in the supplementary video.

The second part of the experiments highlights the capability of our approach to compensate when provided with a misspecified model. This is shown both by the plots in~\cref{fig:planar_biased_plots} for the planar manipulator and in~\cref{tab:panda_exp} for the Panda robot. 
In the planar case, for no bias, SDLS outperforms \crisp{}-FK. However, just a small bias makes the prediction error of SDLS increase sharply, while \crisp{}-FK shows much higher robustness even for biases up to $10~deg$ in the joint angles or $\sim2~cm$ in the link lengths. 
A similar trend is observed in the experiments with the Panda robot in \cref{tab:panda_exp}. 
In the simpler scenario of the circular trajectory with known orientations, \crisp{}-FK outperforms from the start SDLS. 
On $C_0$ and $S$ an error of half a degree is enough to observe a significant increase in the error for SDLS, while \crisp{}-FK remains more robust.

\subsection{Computational Times}

We make a note on computational times. For reference, all experiments were performed on an Intel\textregistered~Core\texttrademark~i7-1075H laptop CPU. Within this setting, CRiSP requires on average $5$ minutes to train (and perform model selection) on $25000$ training points and $0.7$s for each prediction.
SDLS (which is model-based and does not have a training phase) takes $10^{-4}$s on average at prediction time. 
% In comparison, OCSSVM requires $12$ minutes for training and $0.001$s for testing. 

As expected, the SDLS model-based approach is significantly faster than the data-driven strategies.
% \raf{Still, it does not provide strict guarantees on joint limits avoidance, as CRiSP instead does by solving a more computationally demanding constrained optimization problem.}
However, the implementation of SDLS, albeit using a Python interface, is based on the C++ layer of the Bullet simulator, while CRiSP is completely in Python, except for the matrix inversion routine used in the training step. 
We note that these numbers are highly dependant on the complexity of the forward kinematics and the number of training points. While the focus of our work is on the robustness and predictive accuracy of CRiSP, we argue that further work can significantly improve prediction time by leveraging classical large-scale machine learning techniques such as Nystr\"om approximation, determinantal point sampling or gradient preconditioning.
% Therefore, there is still large room to improve the computational of the proposed approach, potentially reducing the gap with model-based predictions by a significant amount. 

% \begin{figure}[]
%     \centering
%     \begin{subfigure}[]{.3\textwidth}
%         \centering
%     	\includegraphics[width=\textwidth,trim={0.2cm 2cm 0.1cm 1.3cm},clip]{figures/pos_tracking_spiral_bias_1e-1.png}
%     	\caption{}
%         \label{fig:spiral_tracking}
%     \end{subfigure}
%     ~
%     % \hspace{-3em}
%     \begin{subfigure}[]{.19\textwidth}
%         \centering
%         \includegraphics[width=\textwidth]{circle_2.png}
%         \caption{}
%         \label{fig:circ_tracking}
%     \end{subfigure}
%     \caption{\small
%      Trajectories reconstructed by CRiSP-FK and SDLS under model bias (0.1 $deg$) for the $S$ (a) and $C_0$ (b) tasks, respectively.}
%   \label{fig:pandatraj}
% \end{figure}

%===============================================================================

\section{Conclusions}
\label{sec:conclusion}

In this work we studied the problem of learning the inverse kinematics of a robot manipulator using a data-driven approach. We focused on settings in which the kinematic structure of the robot is known, but potentially inaccurate. Within this setting, we proposed \crisp{}, a structured prediction algorithm combining a-priori knowledge of the model with non-parametric regression, to efficiently learn an inverse kinematics map. We characterized the generalization properties of the proposed approach and empirically demonstrated the effectiveness of \crisp{} on trajectory reconstruction tasks. Our approach is significantly more effective than previous data-driven methods, as well as model-based ones, even in settings affected by varying types of bias in the kinematics.

Future research will focus on two main directions: firstly, we will explore acceleration strategies at the inference stage, to make our method appealing for real-time applications. 
Secondly, we will investigate active-learning-based strategies to find better anchor points to train \crisp{}, ultimately yielding a more concise and efficient estimator.

% {\color{red}We studied the problem of IK in joint space showing its limits. We propose a novel approach to the problem that is data-driven but can use the information of a biased analytic model showing robustness to errors in the model. We tested it on two scenario with promising results.
% Future directions: acceleration (Nystrom, Leverage scores), AL for finding the best points, RL?}

%%%%%%%%%%%%%%%%%%%%%%%%%%%%%%%%%%%%%%%%%%%%%%%%%%%%%%%%%%%%%%%%%%%%%%%%%%%%%%%%

%%%%%%%%%%%%%%%%%%%%%%%%%%%%%%%%%%%%%%%%%%%%%%%%%%%%%%%%%%%%%%%%%%%%%%%%%%%%%%%%



\section*{Acknowledgements}

This material is based upon work supported by the Center for Brains, Minds and Machines (CBMM), funded by NSF STC award CCF-1231216, and the Italian Institute of Technology. We gratefully acknowledge the support of NVIDIA Corporation for the donation of the Titan Xp GPUs and the Tesla k40 GPU used for this research.
Part of this work has been carried out at the Machine Learning Genoa (MaLGa) center, Università di Genova (IT).
We thank Silvio Traversaro and Yeshasvi Tirupachuri from Istituto Italiano di Tecnologia for the fruitful discussions and insights.
L. R. acknowledges the financial support of the European Research Council (grant SLING 819789), the AFOSR projects FA9550-18-1-7009, FA9550-17-1-0390 and BAA-AFRL-AFOSR-2016-0007 (European Office of Aerospace Research and Development), and the EU H2020-MSCA-RISE project NoMADS - DLV-777826. C.C. acknowledges the financial support of the Royal Society of Engineering, grant SPREM RGS/R1/201149.

\bibliographystyle{plain}
\bibliography{biblio}

\begin{thebibliography}{10}

\bibitem{adams2003sobolev}
Robert~A Adams and John~JF Fournier.
\newblock {\em Sobolev spaces}.
\newblock Elsevier, 2003.

\bibitem{bakir2007predicting}
GH~Bakir, T~Hofmann, B~Sch{\"o}lkopf, AJ~Smola, B~Taskar, and SVN Vishwanathan.
\newblock Predicting structured data, 2007.

\bibitem{bocsi2011learning}
Botond B{\'o}csi, Duy Nguyen-Tuong, Lehel Csat{\'o}, Bernhard Schoelkopf, and
  Jan Peters.
\newblock Learning inverse kinematics with structured prediction.
\newblock In {\em International Conference on Intelligent Robots and Systems}.
  IEEE, 2011.

\bibitem{buss2005selectively}
Samuel~R Buss and Jin-Su Kim.
\newblock Selectively damped least squares for inverse kinematics.
\newblock {\em Journal of Graphics tools}, 10(3):37--49, 2005.

\bibitem{byrd1995limited}
Richard~H Byrd, Peihuang Lu, Jorge Nocedal, and Ciyou Zhu.
\newblock A limited memory algorithm for bound constrained optimization.
\newblock {\em SIAM Journal on Scientific Computing}, 16(5):1190--1208, 1995.

\bibitem{ciliberto2016consistent}
Carlo Ciliberto, Lorenzo Rosasco, and Alessandro Rudi.
\newblock A consistent regularization approach for structured prediction.
\newblock In {\em Advances in neural information processing systems}, pages
  4412--4420, 2016.

\bibitem{ciliberto2020general}
Carlo Ciliberto, Lorenzo Rosasco, and Alessandro Rudi.
\newblock A general framework for consistent structured prediction with
  implicit loss embeddings.
\newblock {\em Journal of Machine Learning Research}, 21(98):1--67, 2020.

\bibitem{ciliberto2017consistent}
Carlo Ciliberto, Alessandro Rudi, Lorenzo Rosasco, and Massimiliano Pontil.
\newblock Consistent multitask learning with nonlinear output relations.
\newblock In {\em Advances in Neural Information Processing Systems}, pages
  1986--1996, 2017.

\bibitem{coumans2016pybullet}
Erwin Coumans and Yunfei Bai.
\newblock Pybullet, a python module for physics simulation for games, robotics
  and machine learning.
\newblock 2016.

\bibitem{de2008learning}
Vicente~Ruiz De~Angulo and Carme Torras.
\newblock Learning inverse kinematics: Reduced sampling through decomposition
  into virtual robots.
\newblock {\em IEEE Transactions on Systems, Man, and Cybernetics, Part B
  (Cybernetics)}, 2008.

\bibitem{d2001learning}
Aaron D'Souza, Sethu Vijayakumar, and Stefan Schaal.
\newblock Learning inverse kinematics.
\newblock In {\em Proceedings 2001 IEEE/RSJ International Conference on
  Intelligent Robots and Systems.}, volume~1, pages 298--303. IEEE, 2001.

\bibitem{george2018control}
Thomas George~Thuruthel, Yasmin Ansari, Egidio Falotico, and Cecilia Laschi.
\newblock Control strategies for soft robotic manipulators: A survey.
\newblock {\em Soft robotics}, 2018.

\bibitem{joachims2009predicting}
Thorsten Joachims, Thomas Hofmann, Yisong Yue, and Chun-Nam Yu.
\newblock Predicting structured objects with support vector machines.
\newblock {\em Communications of the ACM}, 52(11):97--104, 2009.

\bibitem{jordan1992forward}
Michael~I Jordan and David~E Rumelhart.
\newblock Forward models: Supervised learning with a distal teacher.
\newblock {\em Cognitive science}, 1992.

\bibitem{nowozin2011structured}
Sebastian Nowozin, Christoph~H Lampert, et~al.
\newblock Structured learning and prediction in computer vision.
\newblock {\em Foundations and Trends{\textregistered} in Computer Graphics and
  Vision}, 6(3--4):185--365, 2011.

\bibitem{oyama2001inverse}
Eimei Oyama, Nak~Young Chong, Arvin Agah, and Taro Maeda.
\newblock Inverse kinematics learning by modular architecture neural networks
  with performance prediction networks.
\newblock In {\em Proceedings 2001 ICRA. IEEE International Conference on
  Robotics and Automation}, 2001.

\bibitem{oyama2005inverse}
Eimei Oyama and Taro et~al. Maeda.
\newblock Inverse kinematics learning for robotic arms with fewer degrees of
  freedom by modular neural network systems.
\newblock In {\em 2005 IEEE/RSJ International Conference on Intelligent Robots
  and Systems}, 2005.

\bibitem{rudi2018manifold}
Alessandro Rudi, Carlo Ciliberto, Gian~Maria Marconi, and Lorenzo Rosasco.
\newblock Manifold structured prediction.
\newblock In {\em Advances in Neural Information Processing Systems}, pages
  5610--5621, 2018.

\bibitem{siciliano1990kinematic}
Bruno Siciliano.
\newblock Kinematic control of redundant robot manipulators: A tutorial.
\newblock {\em Journal of intelligent and robotic systems}, 1990.

\bibitem{spong2006robot}
Mark~W Spong, Seth Hutchinson, Mathukumalli Vidyasagar, et~al.
\newblock {\em Robot modeling and control}, volume~3.
\newblock wiley New York, 2006.

\bibitem{steinwart2008support}
Ingo Steinwart and Andreas Christmann.
\newblock {\em Support vector machines}.
\newblock Springer, 2008.

\bibitem{takeuchi2006nonparametric}
Ichiro Takeuchi, Quoc~V Le, Timothy~D Sears, and Alexander~J Smola.
\newblock Nonparametric quantile estimation.
\newblock {\em Journal of machine learning research}, 2006.

\bibitem{tevatia2000inverse}
Gaurav Tevatia and Stefan Schaal.
\newblock Inverse kinematics for humanoid robots.
\newblock In {\em Proceedings 2000 ICRA. IEEE International Conference on
  Robotics and Automation.}, volume~1, pages 294--299. IEEE, 2000.

\bibitem{virtanen2020scipy}
Pauli Virtanen, Ralf Gommers, et~al.
\newblock Scipy 1.0: fundamental algorithms for scientific computing in python.
\newblock {\em Nature methods}, 2020.

\end{thebibliography}

\end{document}